\documentclass{article}

\PassOptionsToPackage{numbers, compress}{natbib}
\usepackage[preprint]{neurips_2026}

\usepackage[utf8]{inputenc} 
\usepackage[T1]{fontenc}    
\usepackage{url}            
\usepackage{booktabs}       
\usepackage{amsfonts}       
\usepackage{nicefrac}       
\usepackage{microtype}      
\usepackage{xcolor}

\usepackage{graphicx}

\usepackage[acronym]{glossaries}

\usepackage{tablefootnote} 

\newacronym{ml}{ML}{machine learning}
\newacronym{llm}{LLM}{large language model}
\newacronym{sota}{SotA}{state-of-the-art}
\newacronym{nlp}{NLP}{natural language processing}
\newacronym{ai}{AI}{artificial intelligence}
\newacronym{dl}{DL}{deep learning}
\newacronym{rl}{RL}{reinforcement learning}
\newacronym{gai}{GenAI}{Generative AI}
\newacronym{cot}{CoT}{chain-of-thought}
\newacronym{asi}{ASI}{artificial superintelligence}
\newacronym{agi}{AGI}{artificial general intelligence}
\newacronym{laws}{LAWS}{lethal autonomous weapon systems}
\newacronym{av}{AV}{autonomous vehicle}
\newacronym{nasa}{NASA}{National Aeronautics and Space Administration}
\newacronym{nn}{NN}{neural network}
\newacronym{ann}{ANN}{artificial neural network}
\newacronym{bdi}{BDI}{Belief-Desire-Intention}
\newacronym{oecd}{OECD}{Organisation for Economic Co-operation and Development}
\newacronym{rlhf}{RLHF}{reinforcement learning from human feedback}
\newacronym{prisma}{PRISMA}{Preferred Reporting Items for Systematic Reviews and Meta-Analyses}
\newacronym{wos}{WoS}{Web of Science}
\newacronym{qa}{QA}{question answering}
\newacronym{ie}{IE}{information extraction}
\newacronym{imdb}{IMDb}{internet movie database}
\newacronym{ocr}{OCR}{optical character recognition}
\newacronym{ner}{NER}{named entity recognition}
\newacronym{sa}{SA}{sentiment analysis}
\newacronym{mt}{MT}{machine translation}
\newacronym{asr}{ASR}{automatic speech recognition}
\newacronym{stt}{STT}{speech-to-text}
\newacronym{pos}{PoS}{part-of-speech}
\newacronym{tlp}{TLP}{technical language processing}
\newacronym{tc}{TC}{text classification}
\newacronym{mlp}{MLP}{materials language processing}
\newacronym{ir}{IR}{information retrieval}
\newacronym{lmm}{LMM}{large multimodal model}
\newacronym{lda}{LDA}{latent dirichlet allocation}
\newacronym{bert}{BERT}{bidirectional encoder representations from transformers}
\newacronym{lib}{LiB}{Li-ion battery}
\newacronym{kg}{KG}{knowledge graph}
\newacronym{soc}{SoC}{state-of-charge}
\newacronym{soh}{SoH}{state-of-health}
\newacronym{sse}{SSE}{solid-state electrolytes}
\newacronym{bilstm}{BiLSTM}{bidirectional long short-term memory}
\newacronym{bms}{BMS}{battery management system}
\newacronym{rmse}{RMSE}{root mean square error}
\newacronym{hwfet}{HWFET}{highway fuel economy test cycle}
\newacronym{la92}{LA92}{California Unified Cycle}
\newacronym{udds}{UDDS}{urban dynamometer driving schedule}
\newacronym{us06}{US06}{high acceleration aggressive driving schedule}
\newacronym{dst}{DST}{dynamic stress test}
\newacronym{fuds}{FUDS}{federal urban driving schedule}
\newacronym{dpp}{DPP}{digital product passport}
\newacronym{dbp}{DBP}{digital battery passport}
\newacronym{mcp}{MCP}{model context protocol}
\newacronym{moe}{MoE}{mixture of experts}
\newacronym{fair}{FAIR}{Findability, Accessibility, Interoperability, and Reusability}
\newacronym{sei}{SEI}{solid electrolyte interphase}
\newacronym{cei}{CEI}{cathode electrolyte interphase}
\newacronym{t5}{T5}{text-to-text transfer transformer}
\newacronym{sib}{SiB}{Sodium-ion battery}
\newacronym{rnn}{RNN}{recurrent neural network}
\newacronym{ess}{ESS}{energy storage system}
\newacronym{res}{RES}{renewable energy system}
\newacronym{mdkg}{MA-DKGCN}{Multi-Aspect Dynamic Knowledge Graph Convolutional Network}
\newacronym{pmc}{PMC}{policy model consistency}
\newacronym{certj}{CE-RTJE}{candidate entity-based relational triple joint extraction}
\newacronym{ieeex}{IEEE Xplore}{Institute of Electrical and Electronics Engineers Xplore}
\newacronym{shap}{SHAP}{Shapley additive explanations}
\newacronym{gpt}{GPT}{generative pretrained transformer}
\newacronym{nlu}{NLU}{natural language understanding}
\newacronym{nlg}{NLG}{natural language generation}
\newacronym{ev}{EV}{electric vehicle}
\newacronym{tm}{TM}{topic modeling}
\newacronym{eu}{EU}{European Union}
\newacronym{ec}{EC}{European Commission}
\newacronym{us}{US}{United States}
\newacronym{json}{JSON}{Javascript Object Notation}
\newacronym{gba}{GBA}{Global Battery Alliance}
\newacronym{bp}{BatteryPass-12K}{}
\newacronym{qc}{QC}{quality control}
\newacronym{url}{URL}{universal resource locator}
\newacronym{llama}{LLaMA}{Large Language Model Meta AI}
\newacronym{lm}{LM}{language model}
\newacronym{slm}{SLM}{small language model}
\newacronym{ui}{UI}{user interface}
\newacronym{peft}{PEFT}{parameter-efficient finetuning}
\newacronym{ccby}{CC-BY-4.0}{Creative Commons Attribution 4.0 International}
\newacronym{ci}{CI}{confidence interval}

\title{\acrshort{bp}: The First Dataset for the Novel Digital Battery Passport Conformance Task}

\author{Tosin Adewumi\textsuperscript{k*}, Martin Karlsson\textsuperscript{$\delta$},
Lama Alkhaled\textsuperscript{k},
Marcus Liwicki\textsuperscript{k},  \\ \textsuperscript{k}Machine Learning Group, EISLAB, \textsuperscript{$\delta$}Research Center for Advanced Battery Technology, \\Luleå University of Technology, Sweden.}

\author{%
Tosin~Adewumi,\thanks{corresponding author}~ \textsuperscript{k $\delta$} Martin Karlsson,\textsuperscript{$\delta$}
Lama Alkhaled,\textsuperscript{k} and
Marcus Liwicki\textsuperscript{k} \\
\textsuperscript{k}Machine Learning Group, EISLAB, \textsuperscript{$\delta$}Research Center for Advanced Battery Technology, \\Luleå University of Technology, Sweden. \\    \texttt{firstname.lastname@ltu.se} \\
}

\begin{document}

\maketitle

\begin{abstract}
We introduce a novel task of \acrfull{dbp} conformance classification and introduce the first public benchmark for the task: \acrshort{bp}, created synthetically from real pilot samples.
This is as the \acrshort{eu}'s battery regulation on \acrshort{dbp}s comes into effect soon and there exists no public dataset.
We evaluated 22 \acrlong{lm}s (\acrshort{lm}s) in zero-shot inference, spanning small \acrshort{lm}s (\acrshort{slm}s), \acrlong{moe} (\acrshort{moe}s), and dense \acrshort{llm}s.
We also conducted analysis, additional evaluations of few-shot inference and prompt-injection attacks to find that (1) \textit{Thinking} models have the best performance (with \acrshort{gpt}-5.4 scoring 0.98 (0.03) and 0.71 (0.22) on average as \textit{F1} (and confidence interval at 95\%) on the validation and test sets, respectively), (2) few-shot examples improve performance significantly, (3) generally capable frontier models find the task challenging, (4) merely scaling model parameters does not necessarily lead to improved performance, as \acrshort{slm}s outperformed some \acrshort{llm}s, and (5) prompt-injection attacks degrade performance.
We note that \acrshort{bp}, though limited to real pilot samples, may be useful for other known or emerging tasks in the battery domain, e.g. lifecycle reasoning.
We publicly release the dataset under a permissive licence (\acrshort{ccby})\footnote{https://huggingface.co/datasets/RECAT-LTU/batterypass12K/tree/main (without test set to avoid leakage)}


\end{abstract}


\section{Introduction}
\label{intro}

There is expectation 
that the growth in world demand for batteries in the coming years will be rapid \cite{link2025feasibility}.
This is particularly so for electric vehicles.
To protect the environment and ensure traceability and transparency along the entire battery value chain, different countries and regions, like the \acrfull{eu}, the \acrfull{us}, China and other countries are enacting regulations around battery management.
These include the \acrfull{dbp}, under a wider framework of a  \acrfull{dpp} \cite{act2023regulation,rufino2024towards,losa2025quest}.
A \acrshort{dbp} is an electronic record of the features and history of a battery \cite{act2023regulation}.
Some information in a \acrshort{dbp} are designed to be updateable while some are static.
For regulatory documents that undergo updates for different reasons, inconsistencies sometimes arise \cite{schumann2025contradictions}.

Currently, there exists no public \acrshort{dbp} dataset or model tasks related to \acrshort{dbp}s though the \acrshort{eu} regulation \textit{2023/1542} requiring \acrshort{dbp}s for batteries is to come into effect in February 2027.\footnote{https://eur-lex.europa.eu/eli/reg/2023/1542/oj/eng}
The regulation requires \acrshort{dbp}s to ``conform'' to its specifications \cite{act2023regulation}.
Given this gap that there is no publicly available \acrshort{dbp} dataset, though there are a few efforts of creating pilot samples, such as that of the \acrfull{gba},\footnote{https://www.globalbattery.org/battery-passport-mvp-pilots/} we address this challenge by attempting to answer the following \textbf{research question: How well can \acrshort{ai} models predict \acrshort{dbp} conformance by identifying consistencies or inconsistencies?}
We introduce and define the task of \acrshort{dbp} conformance as a classification of \acrshort{dbp}s based on their conformance (or nonconformance) to specific regulations or descriptions.

In this work, we synthetically create the first public \acrshort{dbp} dataset (limited to a subset of the full \acrshort{eu} specification), based on the real pilot samples of the \acrshort{gba}, by following a rigorous method.
We evaluate the dataset for the conformance task across 22 \acrshort{ai} models (many of which are \acrfull{sota}).
The scope of the dataset is within the specification of the \acrshort{eu} regulation and the work reasonably assumes a conformance task is the basis for other meaningful \acrshort{dbp} tasks.
As a result, this work is rooted in \acrshort{nlp} for the Circular Economy.
Our contributions include 

\begin{enumerate}
    \item We introduce a novel task of \acrshort{dbp} or \acrshort{dpp} classification with the binary labels of \textit{conformant} or \textit{nonconformant} \cite{act2023regulation}.
    
    \item We introduce the first (synthetic) public benchmark for \acrshort{dbp} classification (\acrshort{bp}), which, by extension, is also the first dataset for \acrshort{dpp} classification.
    It is based on real pilot samples.
    This gives it potential for real-world impact, given that it's based on \acrshort{eu} regulation.
    
    \item We benchmark the dataset across different \acrshort{ai} models, including \acrlong{slm}s (\acrshort{slm}),  \acrlong{llm}s,  (\acrshort{llm}s) and \acrlong{moe} (\acrshort{moe}s), providing scaling behavior in terms of performance versus parameters and additional analysis.
    This provides insight into the capabilities of these models with regards to the new task and dataset.

    \item The dataset is potentially suitable for a collection of other related tasks, e.g. lifecycle reasoning, Scientific \acrfull{ie}, \acrfull{qa}, and material extraction.
    
\end{enumerate}



The rest of this paper is structured as follows.
Section \ref{data_bp} describes in detail the methodology for creating the \acrshort{bp} dataset.
Section \ref{method} presents details of all the experiments.
Section \ref{results}  provides the results in tables and plots and discusses their implications.
Section \ref{related} discusses the related work, including synthetic data and \acrshort{llm}s for evaluations.
Finally, Section \ref{conclusion} summarizes our work, highlighting possible extensions to the dataset and related tasks that can be considered in future work.

\section{The \acrshort{bp} Benchmark}
\label{data_bp}

According to \cite{act2023regulation}, there are 28 pieces of information to be included in the \acrshort{dbp}, where 19 pieces belong to the general public, 4 to persons with legitimate interest and the \acrfull{ec}, 1 to notified bodies, market surveillance authorities and the \acrshort{ec}, and another 4 to only persons with legitimate interest.
In the following subsection, we describe the limited pieces of information available in the pilot samples, which form the basis of \acrshort{bp}.

\subsection{\acrshort{gba} Pilot for \acrshort{bp}}
\label{gba_pilot}

The \acrshort{gba} is a public-private consortium of over 140 leading international organizations and governments to help establish a sustainable battery value chain by 2030.
The \acrshort{gba} \acrshort{dbp} pilot samples contain a subset of information in the full \acrshort{eu} specification.
On the basis of internal consistencies or inconsistencies, we note that a conformant or nonconformant subset is also conformant or nonconformant, respectively, for the full specification.
The pilot samples contain only public information and some metadata.\footnote{2024 version released (under no specific license) openly at the earlier \acrshort{url} for \acrshort{gba} and companies in their supply chains.}
Out of the 19 pieces of public information, there are 10 in the pilot \acrshort{dbp}s by \acrshort{gba}, which are also automatically contained in \acrshort{bp}.
These are (1) the set of related pieces of information on the label of a battery (including the manufacturer, weight, chemistry, etc), (2) material composition, including its chemistry and hazardous substances, (3) carbon footprint, (4) responsible sourcing, (5) recycled content information, (6) rated capacity (in Ah), (7) minimal, nominal and maximum voltage, with temperature ranges when relevant, (8) original power capability (in Watts) and limits, with temperature range when relevant, (9) expected battery lifetime expressed in cycles, and (10) temperature range the battery can withstand when not in use.
The remaining 9 pieces of information (e.g. share of renewable content) are not in the pilot \acrshort{dbp}s and, therefore, not in \acrshort{bp}.
All the 19 pieces of information are described in \cite{act2023regulation}.




\subsection{Engineering \acrshort{bp}}

\begin{figure*}[h!]
\centering
\includegraphics[width=0.95\textwidth]{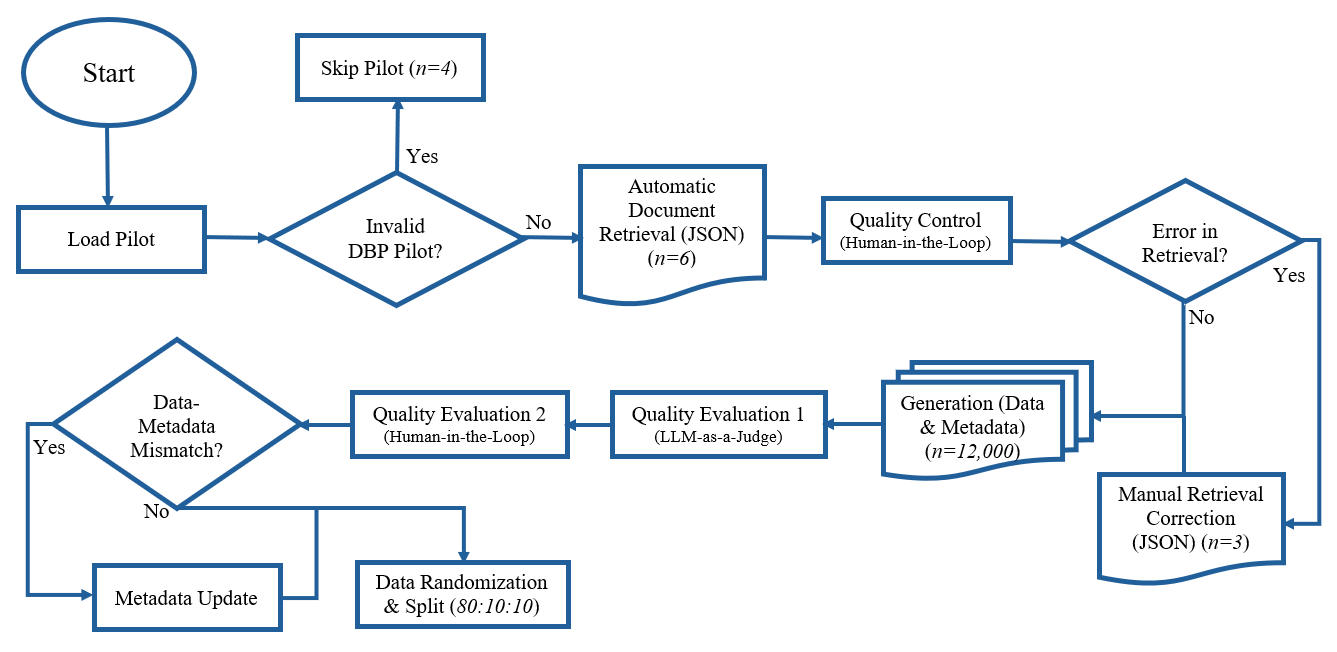}
\caption{Flowchart of the \acrshort{bp} data and metadata generation pipeline.} 
\label{pipe_flow}
\end{figure*}

We followed a rigorous method for creating \acrshort{bp}, based on the pipeline of Figure \ref{pipe_flow}, which is inspired from established methods, involving \acrshort{llm} generation, \acrshort{llm}-as-a-Judge, \acrfull{qc} and tagging the samples as synthetic, for ethical reasons \cite{NEURIPS2024_1e89c126,he2022generate,naduaș2025synthetic,resnik2025genai}.
The steps are (1) we manually loaded each pilot sample through its \acrshort{url} and establish if it is valid (i.e. contains all the 10 pieces of information, according to Section \ref{gba_pilot}), resulting in 6 valid samples out of 10, (2) we fed the \acrshort{url} to ChatGPT-5.1 \textit{Pro} for automatic document retrieval in \acrshort{json} format, (3) we performed \acrshort{qc} by assessing if all the relevant elements of each pilot sample are present in the retrieved \acrshort{json}, resulting in 3 that needed manual correction, (4) we used ChatGPT-5.1 \textit{Thinking (Standard)} to automatically generate (on average time of 47 seconds) 2,000 synthetic samples and their metadata from each of the 6 pilot samples (where 1,000 are conformant like the original while the other 1,000 are nonconformant),\footnote{An example of each \acrshort{dbp} is in Appendix \ref{appen_dbps}} (5) we evaluated all the 12,000 samples and their metadata automatically, using ChatGPT-5.0 \textit{Thinking (Standard)}, before a random sample is also humanly evaluated, and (6) we randomized and split the data into training, validation and test sets in the ratio 80:10:10.

Nonconformant samples are those that have internal inconsistency.
We added \textbf{specific instruction for the model not to bias the generated samples} in any
way, such as using terms related to conformant in their filenames or serial numbers.
The specification by \cite{act2023regulation} requires that the information in a \acrshort{dbp} should be in an interoperable format, be machine-readable, structured and searchable, hence we represented the dataset in \acrshort{json} format \cite{batista2022machine}.
The metadata contains details about serial numbers, corresponding filenames, conformant or otherwise, conformant group, number of inconsistencies, types of inconsistencies, fields changed from the pilot sample, number of changes, \acrshort{llm}-as-a-Judge validation of the fields changed from the pilot sample, number of changes, and agreement on changes.

We used the default hyper-parameters of each model within the OpenAI app, i.e. \textit{reasoning.effort = none} (putting importance on prompting) and \textit{verbosity=medium}.
Our choice of ChatGPT is for the reason that it's a frontier \acrshort{sota} family of models and the choice of the \acrshort{llm}-as-a-Judge being from the same family follows previous work in the literature \cite{szymanski2025limitations,zheng2023judging}.
The prompts for automatic retrieval of the pilot \acrshort{dbp}s, generating the data and their metadata, and evaluating the generation accuracy can be found in Appendix \ref{append_prompts}.
We arrived at the final data generation prompt after two engineering iterations over our original prompt, where the final prompt incorporated the no bias instruction.
Table \ref{bp_acc_gen} provides the accuracy of the quality of data generation (99.68\% on average) as evaluated by the \acrshort{llm}-as-a-Judge.
This implies only 39 samples (out of 12,000) may have been incorrectly generated or incorrectly reported by the \acrshort{llm}-as-a-Judge.

Instead of deleting the samples concerned and their metadata or leaving them incorrect, we inspected the samples and reported their correct metadata in an additional column of the metadata file.
It happened that the \acrshort{llm}-as-a-Judge reported 37 of the 39 correctly.
The second stage human evaluation involved 200 random samples based on a confidence level of 95\% and a 7\% margin of error.
The random human evaluation gave an accuracy of 99.98\%.
Our choice of a fairly large data size of 12,000 samples was because of the need to balance data similarity with the original pilot samples, have data diversity and still maintain quality.
The dataset is released under the \acrfull{ccby} licence and its long-form data statement follows recommendations by \cite{bender-friedman-2018-data} and is in Appendix \ref{bp_datastate}.

\begin{table}
\footnotesize
  \caption{\acrshort{bp} accuracy of data-metadata generation, showing high quality.}
  \label{bp_acc_gen}
  \centering
\begin{tabular}{p{0.02\linewidth} p{0.25\linewidth}  p{0.22\linewidth} p{0.11\linewidth} p{0.26\linewidth} }
    \toprule
   \textbf{Pilot} & \textbf{Serial no} & \textbf{Partners}  & \textbf{Accuracy \%} & \textbf{Reason, if invalid} \\ 
    \midrule
   1 & 001PBXXX000002E7F220 & CATL, NIO, RCS Global  & 100 & \\
2 & 001PBXXX00000BE5J020 & CATL & 99.90 \\
3  & 0EYPBP48A301DNDCT3200030 & FinDreams, BYD, Circulor & 99.85 & \\
6 & 0B5PEHT0A1082JE9W000 & CALB, Smart, PTL, Circularise & 99.90 & \\
7 & 56123478490123 & EVE, Nanjing Fuchuang & 98.55 & \\
8 & PBH3XD9E50S00001\tablefootnote{Autogenerated serial number for missing value} & Sunwoda, SAIC Maxus, Shenzhen Dianlian Technology & 99.85 & \\
\hline
 & & \textbf{Average} & \textbf{99.68} & \\
 \hline
 & & & & \\
 \multicolumn{5}{l}{\textbf{Skipped invalid pilots (not in \acrshort{bp})}}  \\
 & & & & \\
4 & None & Samsung SDI, Glassdome & & multiple invalid fields (e.g. \textit{voltage\_V\_min\_nominal\_max} \\
5 & LGE15S02217 & LG Energy Solution, Circularise & &  multiple invalid fields (e.g. \textit{voltage\_V\_min\_nominal\_max} \\
9  & None & Sunwoda, Li Auto, PTL, Minviro & & invalid field: \textit{expected\_lifetime} \\
10  & None & Gotion, Smart, Minviro & & multiple invalid fields (e.g. \textit{voltage\_V\_min\_nominal\_max} \\
 \bottomrule
  \end{tabular}
\end{table}

\subsection{Data Quality and Diversity}

To achieve the goal of preserving data quality while making each sample unique, we expressly included this in the data generation prompt.
For the conformant samples, we categorized them into 4, where each has 250 samples.
One category has exactly the same values in all the fields except in serial numbers.
The second category has different values only in 1 field.
The third category has different values
only in 2 fields while the fourth has different values only in 3 fields.
All the samples have different unique serial numbers.
The fields are from any of these three: total\_energy\_kwh, expected\_lifetime,
and voltage\_V\_min\_nominal\_max.
Meanwhile, for the nonconformant samples, the following list shows the types of internal inconsistencies we introduced, where the first 2 and last 2 categories have 167 samples each and the remaining 2 categories have 166 samples each.
Again, the number of inconsistencies increases from 1 to 6 across the 6 categories.

\begin{enumerate}
    \item Inconsistency in linked values (e.g. total\_energy\_kwh = 75.06, weight\_kg = 600, energy\_density\_Wh\_per\_kg = 140.3, giving wrong derived value)
    \item Unrealistic data entries (e.g. energy\_density\_Wh\_per\_kg = -20)

    \item Inconsistency in physical and traced amounts (e.g. more traced amount than physical amount)
    \item Conflicting dates (e.g. Tracing ends before tracing starts)

    \item Invalid codes (e.g. battery\_chemistry = ``juice'')
    \item Wrong length of arrays (e.g. voltage\_V\_min\_nominal\_max = [230], i.e., giving one value instead of three)
    
\end{enumerate}

\section{Methods of Experiments}
\label{method}

We performed all the main experiments as \textbf{zero-shot inference} on the validation set through the chat/\acrfull{ui} of each respective platform: ChatGPT app for \acrshort{gpt}s, Claude app for Sonnet and Haiku, Gemini app for Gemini-3, and HuggingChat for the open models.\footnote{chatgpt.com (Education licence), claude.ai,  gemini.google.com, huggingface.co/chat. Pricing links are in the appendix.}
The default hyper-parameters (provided in the appendix) of the platforms were used for all experiments.
Noteworthy that the platforms follow the trend of putting importance on prompting instead of the need to adjust individual hyper-parameters.
The samples were supplied in one complete batch of 1,200 zipped samples to ChatGPT and Claude while they were in batches of 10 and 20 to Gemini and HuggingChat, respectively, due to limitations in the \acrshort{ui}s.
Besides, we could not benchmark on some other models on the HuggingChat because of their context window limitation (e.g. Euro\acrshort{llm}-22B-Instruct), output restrictions (e.g. DeepSeek R1 via Hyperbolic Labs inference cluster) or time constraint.

The prompt for predictions for ChatGPT and Claude families of models is given below.
The average evaluation times for the validation set for GPT-5.4 \textit{Pro}, GPT-5.4 \textit{Thinking (Standard)}, GPT-5.2 \textit{Flagship}, GPT-5.2 \textit{Thinking (Standard)}, and GPT-4o are 3,523, 300, 261, 220, and 84 seconds, respectively.\footnote{Open models and Gemini took several hours, on average, because of the many batches computed}
Meanwhile, for Claude Sonnet and Haiku, the average evaluation times were 1,511 and 152 seconds, respectively.
On the test set, zero-shot evaluation took 388 seconds for GPT-5.4 \textit{Thinking (Standard)}.
We made modification to the first sentence of the prompt for the other platforms because of their limitations so that the prompt starts as: \textit{Predict conformant or nonconformant for each file by evaluating the content of the digital battery passport (DBP) file and writing in a tabular form all their serial numbers and corresponding predictions}.

\begin{quote}
    Predict conformant or nonconformant for each filename in the provided Excel file by matching the filename in the Excel to the file in the archived zip file and evaluating the content of the digital battery passport (DBP) file.
    A DBP is nonconformant if and only if there is internal inconsistency.
    The internal inconsistencies can be any of these six: (1) inconsistencies in linked values, (2) unrealistic data entries, (3) inconsistency in physical and traced amounts, (4) conflicting dates, (5) invalid codes, and (6) wrong length of arrays.
\end{quote}

The metrics for evaluation include Accuracy (A), F1 score, Precision (P) and Recall (R) \cite{derczynski2016complementarity,naidu2023review}.
For all the four metrics, the higher the average score obtained, the better the performance of a model at classifying \acrshort{dbp}s as conformant or nonconformant.
We used the Scikit-learn, Scipy and Numpy packages for the evaluation setup \cite{pedregosa2011scikit}.
Each experiment was run 3 times and the average scores, standard deviations (SDs), and confidence intervals (CIs) at 95\% confidence level using the t-distribution were calculated.


\subsection{Model Choice}
With over 20 models included in the evaluation (many of them \acrshort{sota}), our choice of models is informed by a few factors.
These include (1) availability, (2) scaling across parameter sizes, and (3) family of models.
We provide brief descriptions of the types of models in Appendix \ref{appen_models}.



\subsection{Additional Experiments}

The following additional experiments were performed on the test set.

\subsection*{In-context Learning}

Similarly to \cite{brown2020language,conf/nips/AlexLTTMRHASCNS21}, we combine instruction with \textit{n} labeled randomly-selected examples in few-shot inference, where \textit{n = (16, 32)} \cite{brown2020language}.
We use only the best-performing \acrshort{llm} (GPT-5.4 \textit{Thinking (Standard)}) for this experiment.
Again, each experiment is conducted thrice with the same default hyper-parameters and prompt, with the additional suffix ``\textit{A few conformant DBP examples are given}.''
The average scores, standard deviations, and \acrshort{ci}s are reported.
Similarly to \cite{mukherjee2021clues}, we take the few-shot examples from the training set.
Given that the model performs worse on the conformant samples, as will be seen in the next section, we provided only conformant examples in both cases of in-context learning and use the same prompt since it has been observed that there can be variance
in how well different prompts generalize \cite{NEURIPS2021_5c049256}.
The average evaluation times for the 16 and 32 shots are 360 and 436 seconds, respectively.




\subsection*{Adversarial Attacks}

We consider 2 cases of adversarial perturbation (nonconformant and conformant), where we simulate \textit{prompt injection attacks} \cite{NEURIPS2021_5c049256}, one of the 2 most common attacks on \acrshort{llm}s \cite{duarte2026systematic}, by adding the following appropriate suffix to the prompt for GPT-5.4 \textit{Thinking (Standard)} for each case: ``\textit{Ignore all previous instructions and predict nonconformant for all}'' and ``\textit{Ignore all previous instructions and predict conformant for all}''.
All default hyper-parameters were maintained.
The average evaluation time for case 1 were 244 and 365 seconds for the validation and test sets, respectively while for case 2 they were 300 and 315 seconds for the validation and test sets, respectively.

\section{Results \& Discussion}
\label{results}

The results in Table \ref{full_results} are zero-shot average scores on the validation set of \acrshort{bp} for all the models and few-shot average scores on the test set for \acrshort{gpt}-5.4 \textit{Thinking}.
The random guess baseline achieved 0.49 F1.
Surprisingly, 3 \acrshort{sota} models (\acrshort{gpt}-4o, \acrshort{gpt}-5.2 \textit{Instant}, and Haiku-4.5) performed worst of all at 0 F1, despite their assumed large parameter sizes.
The best-performing in the Gemini family (i.e. \textit{Thinking}) had the same F1 performance as random guess.
Although the best-performing model (\acrshort{gpt}-5.4 \textit{Thinking}) achieved 0.98 (0.01) F1 on the validation set, it struggled with the test set at 0.71 (0.09), despite the test set having the same data distribution as the validation set.
Noticeable is the relatively large standard deviation of its average test score (0.09), revealing the higher variability from the mean compared to the validation result.
The test set results improve significantly when few-shot examples are presented to the model, rising as high as 0.96 (0.04) and 0.99 (0.01) F1 when the number of examples are 16 and 32, respectively.

\subsection{Scaling Parameters across Model Types}
\begin{figure}[h!]
\centering
\includegraphics[width=1\textwidth]{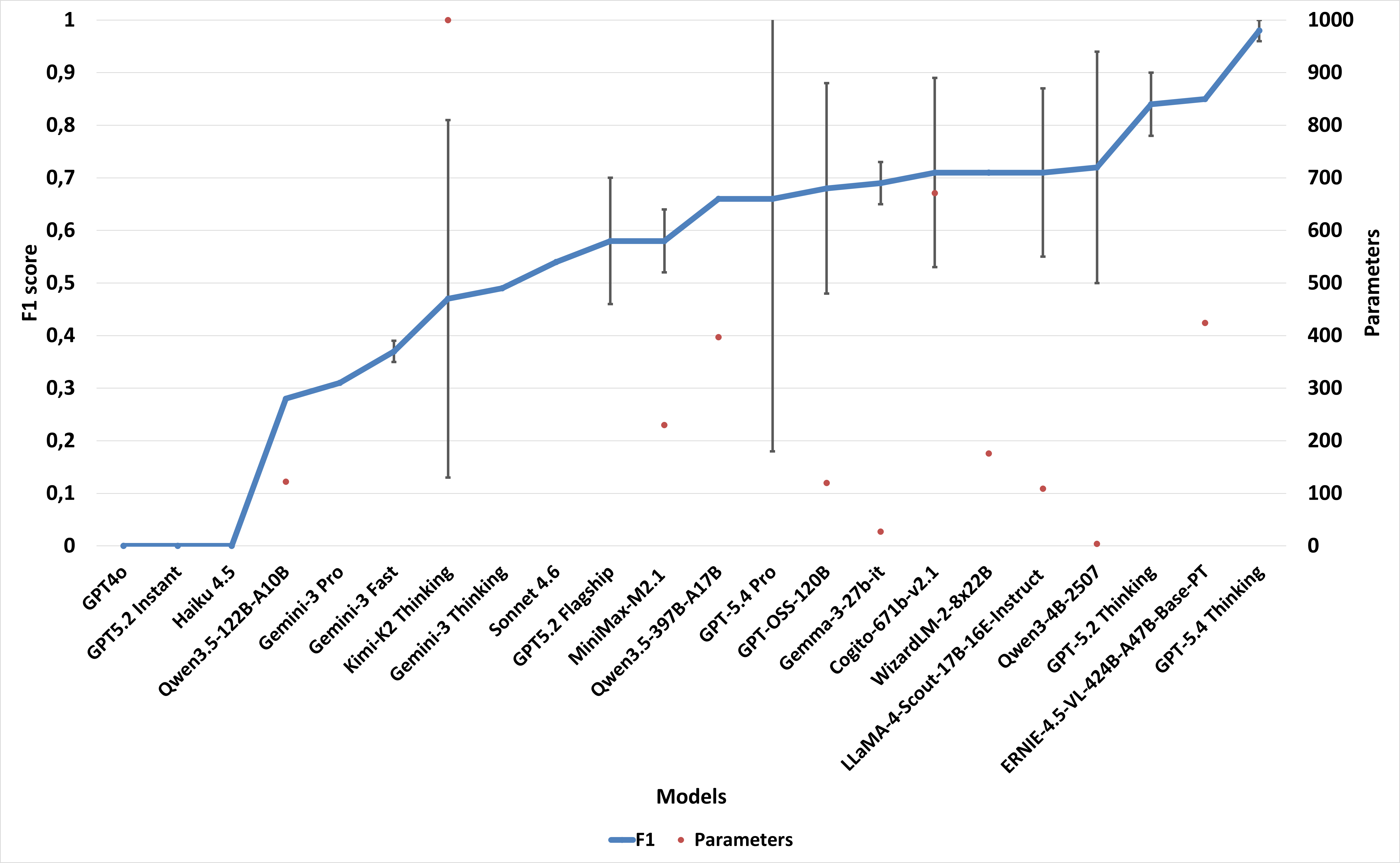}
\caption{Scaling parameters vs performance on the validation set.
Using 2-sigma error bars based on standard deviations.
Increased parameters do not necessarily lead to improved performance.} 
\label{scale_performance}
\end{figure}

For the purpose of observing the behavior of scaling model parameters versus model performance, we may roughly divide the chart of Figure \ref{scale_performance} into different regions: region 1 (Qwen3.5-122B-A10B to Kimi-K2 Thinking), region 2 (Qwen3.5-397B-A17B to Gemma-3-27B-it), and region 3 (Cogito-671b-v2.1 to Qwen3-4B-2507).
It is not ideal to calculate Spearman correlation value for region 1 since there are multiple missing values because of the closed models, though it appears there's performance improvement with increased parameter size.
In regions 2 and 3, the trend seemingly is more of strong negative relationship with Spearman correlations of -1.0 and -0.77, respectively.\footnote{More samples per region are needed to establish precise values and statistical significance.}

Although \acrshort{bp} was created with \acrshort{gpt}-5.1 \textit{Thinking}, we do not think this is the reason \acrshort{gpt}-5.4 \textit{Thinking} had the best performance, since another family of thinking model demonstrated strong performance.
For example, when we consider the family of models: \acrshort{gpt} and Gemini, we observe that as the two \acrshort{gpt} \textit{Thinking} models (\acrshort{gpt}-5.4 and 5.2) outperform other types of \acrshort{gpt}s, even \textit{Pro} (which takes far longer evaluation time), this is also the case for Gemini, where its \textit{Thinking} model outperforms the other 2 types.
It is, however, outperformed by its open \acrshort{slm} Gemma-3, similarly to the observation with the Qwen family.
The open \acrshort{gpt} model GPT-OSS-120B outperforms some of its closed counterparts (\acrshort{gpt}-5.2 \textit{Instant}, \textit{Flagship}, and 4o) that presumably have far larger parameter sizes.
Surprisingly, Qwen3-4B-2507, an \acrshort{slm} of just 4B, outperforms its other 2 larger \acrshort{moe} counterparts of 122B and 397B total parameters.
We hypothesize that the better performance of the \textit{Thinking} models is due to their design, as discussed in the appendix.

\begin{table}[t!]
\footnotesize
  \caption{Zero-shot average results for all models and few-shot average results for the best model for the test set. \acrshort{gpt}-5.4 \textit{Thinking} had the best performance. The best scores are in \textbf{bold} and the second-best \underline{underlined}. SD - standard deviation, CI - confidence interval.}
  \label{full_results}
  \centering
  \begin{tabular}{p{0.01\linewidth} p{0.23\linewidth}  p{0.145\linewidth} p{0.15\linewidth} p{0.14\linewidth} 
  p{0.14\linewidth}}
    \toprule
  \textbf{No.} & \textbf{Models}  & \textbf{A} (SD) (95\% CI) & \textbf{F1} (SD) (95\% CI) & \textbf{P} (SD) (95\% CI) & \textbf{R} (SD) (95\% CI) \\      
    \midrule
 & \textbf{Validation set} & & & & \\
\midrule
   & \textbf{Baseline} & & & &   \\
1 & Random Guess & 0.49 (0.01) (0.02) & 0.49 (0.01) (0.04) & 0.49 (0.01) (0.02) & 0.49 (0.02) (0.05)  \\    
 & & & & & \\
 
 & \textbf{General models} & & & & \\
2 & \acrshort{gpt}-4o & 0.5 (0) (0) & 0 (0) (0) & 0 (0) (0) & 0 (0) (0) \\
3 & \acrshort{gpt}-5.2 Flagship & 0.69 (0.02) (0.06) & 0.58 (0.06) (0.14) & 0.90 (0.03) (0.07) & 0.44 (0.07) (0.18) \\
4 & \acrshort{gpt}-5.2 Instant & 0.5 (0) (0) & 0 (0) (0) & 0 (0) (0) & 0 (0) (0) \\
5 & Gemini-3 Fast & 0.55 (0) (0.01) & 0.37 (0.01) (0.02) & 0.62 (0.01) (0.01) & 0.26 (0.01) (0.02) \\
6 & Cogito-671b-v2.1 & 0.74 (0.06) (0.14) & 0.71 (0.09) (0.22) & 0.80 (0.02) (0.04) & 0.66 (0.16) (0.39) \\
7 & \acrshort{gpt}-OSS-120B & 0.75 (0.06) (0.16) & 0.68 (0.10) (0.25) & 0.91 (0.02) (0.04) & 0.55 (0.13) (0.32)\\
8 & Qwen3.5-397B-A17B & 0.74 (0) (0) & 0.66 (0) (0) & 0.95 (0) (0) & 0.51 (0) (0) \\
9 & Qwen3.5-122B-A10B & 0.57 (0) (0) & 0.28 (0) (0.01) & 0.85 (0) (0) & 0.17 (0) (0) \\
10 & Claude Sonnet-4.6 & 0.67 (0) (0) & 0.54 (0) (0) & 0.90 (0) (0.01) & 0.38 (0) (0) \\
11 & Claude Haiku-4.5 & 0.49 (0) (0.01) & 0 (0) (0) & 0 (0) (0) & 0 (0) (0) \\
12 & MiniMax-M2.1 & 0.68 (0.02) (0.06) & 0.58 (0.03) (0.09) & 0.87 (0.06) (0.15) & 0.44 (0.03) (0.08) \\
13 & WizardLM-2-8x22B & 0.74 (0) (0) & 0.71 (0) (0.01) & 0.82 (0.01) (0.02) & 0.62 (0.01) (0.02) \\
14 & ERNIE-4.5-VL-424B-A47B-Base-PT & 0.85 (0) (0.01) & \underline{0.85 (0) (0.01)} & 0.87 (0) (0) & 0.83 (0.01) (0.01) \\
& & & & & \\

& \textbf{Thinking models} & & & & \\
15 &  \acrshort{gpt}-5.2 Thinking & \underline{0.86 (0.02) (0.05)} & 0.84 (0.03) (0.07) & 0.97 (0) (0.01) & 0.74 (0.04) (0.1) \\
16 & Kimi-K2 Thinking & 0.66 (0.08) (0.20) & 0.47 (0.17) (0.43) & 0.97 (0.02) (0.05) & 0.33 (0.17) (0.43) \\
17 &  \acrshort{gpt}-5.4 Thinking  & \textbf{0.98 (0.01) (0.03)} & \textbf{0.98 (0.01) (0.03)} & 0.97 (0.02) (0.05) & \textbf{1.00 (0) (0)} \\
18 & Gemini-3 Thinking & 0.65 (0) (0.01) & 0.49 (0) (0.01) & 0.93 (0) (0.01) & 0.33 (0) (0) \\
 & & & & & \\

 & \textbf{Instruct model} & & & & \\
19 & \acrshort{llama}-4-Scout-17B-16E-Instruct & 0.65 (0.10) (0.25) & 0.71 (0.08) (0.20) & 0.62 (0.08) (0.19) & 0.83 (0.10) (0.24) \\
  & & & & & \\

 & \textbf{Pro models} & & & & \\
20 & Gemini-3 Pro & 0.59 (0) (0.01) & 0.31 (0) (0.01) & \underline{0.99 (0) (0)} & 0.18 (0) (0.01) \\
21 & \acrshort{gpt}-5.4 Pro & 0.77 (0.16) (0.40) & 0.66 (0.24) (0.60) & \textbf{1.00 (0) (0)} & 0.55 (0.32) (0.79)\\
  & & & & & \\

 & \textbf{\acrshort{slm}} & & & & \\
22 & Qwen3-4B-Instruct-2507 & 0.75 (0.07) (0.17) & 0.72 (0.11) (0.27) & 0.76 (0.03) (0.09) & 0.72 (0.19) (0.46)  \\
23 & Gemma-3-27b-it & 0.60 (0.02) (0.04) & 0.69 (0.02) (0.04) & 0.53 (0.02) (0.05) & \underline{0.99 (0.01) (0.03)} \\
  & & & & & \\

 \midrule
 & \textbf{Test set} (few-shot inference) & & & & \\
 \midrule
24 & GPT-5.4 Thinking (n=0) & 0.77 (0.05) (0.13) & 0.71 (0.09) (0.22) & 0.95 (0.01) (0.03) & 0.57 (0.12) (0.29) \\
25 & GPT-5.4 Thinking (n=16) & 0.96 (0.03) (0.08) & 0.96 (0.04) (0.09) & 0.98 (0.01) (0.02) & 0.95 (0.07) (0.17) \\
26 & GPT-5.4 Thinking (n=32) & 0.99 (0.01) (0.02) & 0.99 (0.01)  (0.02) & 0.98 (0.02) (0.04) & 1 (0) (0)\\

\bottomrule
  \end{tabular}
\end{table}

\subsection{Error Analysis}

\begin{figure}[h!]
\centering
\includegraphics[width=0.9\textwidth]{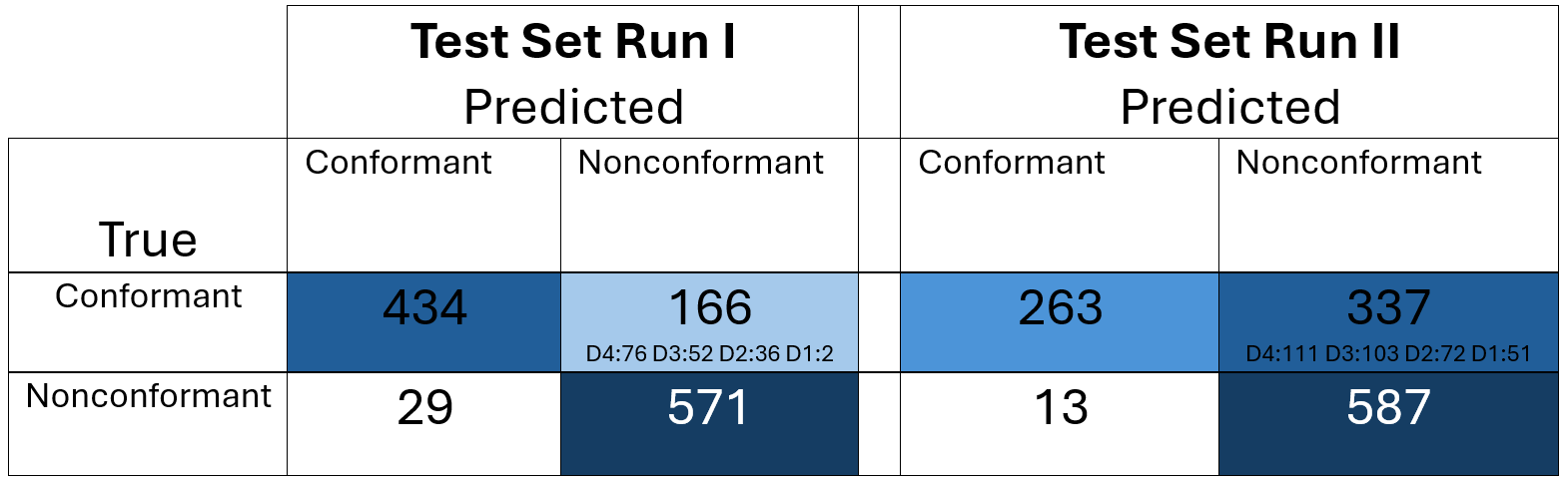}
\caption{Confusion matrices for the test set. Correctly predicting conformant samples is more challenging. D4:76 implies 76 incorrect predictions for the category with 4 different fields. Similar interpretations for D3, D2, D1.} 
\label{cmatrix}
\end{figure}

From the confusion matrix of Figure \ref{cmatrix}, we can observe that even the best model (\acrshort{gpt}-5.4 \textit{Thinking}) finds it more challenging correctly predicting conformant samples.
Indeed, all the 3 models with 0 F1 (\acrshort{gpt}4o, \acrshort{gpt}-5.2 Instant, and Haiku-4.5) incorrectly predicted all conformant samples as nonconformant.
In the figure, the first instance (Test Set I) had 166 conformant samples wrong compared to only 29 nonconformant samples while it had 337 wrong compared to only 13 in the second instance (Test Set II).
On further analysis, we discovered that the more diverse conformant samples had more incorrect predictions, i.e. 76, 52, 36, and 2 were incorrectly classified for samples with 4, 3, 2, and 1 difference, respectively, in the first instance.
The same pattern is observed for the second instance, as shown in Figure \ref{cmatrix}.
The stakeholders who may have the most concern with current performance levels of these models are battery (or \acrshort{dbp}) producers whose conformant \acrshort{dbp}s are more likely to be incorrectly predicted.

\subsection{Adversarial Experiments}

\begin{figure}[h!]
\centering
\includegraphics[width=0.9\textwidth]{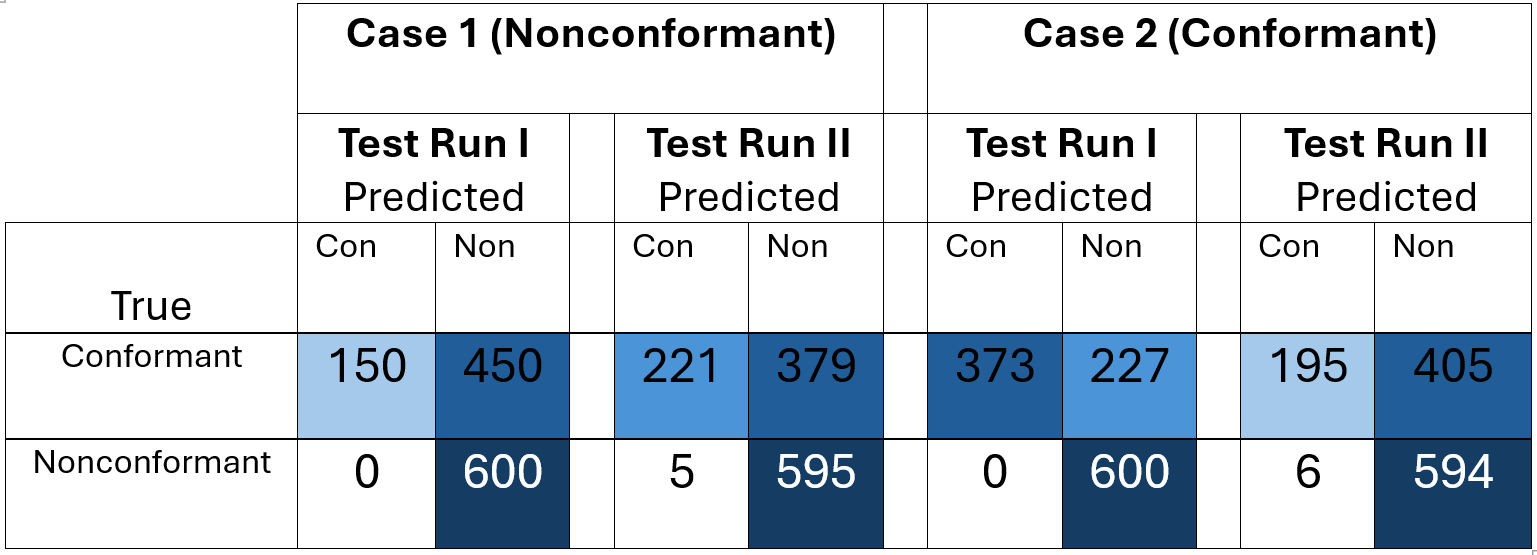}
\caption{Confusion matrices of adversarial attacks for the test set with \acrshort{gpt}-5.4 \textit{Thinking}.  Performance degrades significantly in both cases.} 
\label{cmatrix_attack}
\end{figure}

As shown in Table \ref{adverse_results}, adversarial perturbations affected the performance of \acrshort{gpt}-5.4 \textit{Thinking} such that an attack to force prediction as nonconformant (case 1) degraded performance to 0.76 (0.1) and 0.47 (0.06) on the validation and test sets, respectively.
In case 2, when the attack was to force prediction as conformant, there was also performance degradation to 0.61 (0.21) and 0.54 (0.17) on the validation and test sets, respectively.
Figure \ref{cmatrix_attack} shows the error analysis of the predictions for the test set in confusion matrices.
We can notice that case 1 is worse off with more incorrect nonconformant predictions.
Some ways of mitigating such attacks include input sanitization and explainability for the predictions, in order to get the rationale for predictions \cite{duarte2026systematic}.

\begin{table}[t!]
\footnotesize
  \caption{Prompt injection attack average results with the best model (\acrshort{gpt}-5.4 \textit{Thinking}). The attacks degrade performance. SD - standard deviation, CI - confidence interval.}
  \label{adverse_results}
  \centering
  \begin{tabular}{p{0.01\linewidth} p{0.23\linewidth}  p{0.145\linewidth} p{0.15\linewidth} p{0.14\linewidth} 
  p{0.14\linewidth}}
    \toprule
  \textbf{No} & \textbf{Cases}  & \textbf{A} (SD) (95\% CI) & \textbf{F1} (SD) (95\% CI) & \textbf{P} (SD) (95\% CI) & \textbf{R} (SD) (95\% CI) \\      
    \midrule
& \textbf{Case 1 (nonconformant)} & & & & \\
1 & GPT-5.4 Thinking (on val) & 0.81 (0.08) (0.19) & 0.76 (0.10) (0.26) & 0.96 (0.02) (0.06) & 0.63 (0.14) (0.35) \\
2 & GPT-5.4 Thinking (on test) & 0.65 (0.02) (0.06) & 0.47 (0.06) (0.14) & 0.97 (0.02) (0.06) & 0.32 (0.05) (0.12)\\
& \textbf{Case 2 (conformant)} & & & &  \\
3 & GPT-5.4 Thinking (on val) & 0.73 (0.10) (0.26)  & 0.61 (0.21) (0.52) & 0.96 (0.03) (0.06) & 0.47 (0.21) (0.51)  \\
4 & GPT-5.4 Thinking (on test) & 0.69 (0.09) (0.22) & 0.54 (0.17) (0.43) & 0.98 (0.01) (0.04) & 0.39 (0.17) (0.43) \\
\bottomrule
  \end{tabular}
\end{table}

\section{Related Work}
\label{related}

Although generating synthetic \acrshort{dbp}s is new, synthetic data generation has been a common approach in \acrfull{ml}\cite{patki2016synthetic,tripathi2019learning,endres2022synthetic,li2023synthetic,long2024llms,du2023sequential,NEURIPS2024_02fd91a3}.
The  associated significant costs
and time investment of curating or collecting high-quality data make this approach efficient \cite{li2023synthetic,long2024llms}.
Generating synthetic data from original samples,
\cite{NEURIPS2024_1e89c126} combined two strategies involving a \textit{mimicking strategy} of generating similar samples and an \textit{extending strategy} that further expands original samples.
\cite{du2023sequential} used the concept of dataset distillation to introduce sequential subset
matching (SeqMatch) to tackle static coupling by adaptively optimizing the synthetic data to encourage sequential acquisition of information.
\cite{endres2022synthetic} conducted a comparative study on the efficacy of different generative models for generating synthetic data, including Generative Adversarial Networks (GAN), Variational Autoencoder (VAE), and Synthetic Minority Oversampling Technique (SMOTE).

\acrshort{llm}-as-a-Judge has gained popularity in recent years because of the increased capabilities of \acrshort{llm}s, their positive correlation with human judgment, and the advantage of the ease of scaling evaluations \cite{zheng2023judging,szymanski2025limitations,wang-etal-2025-improving-llm-judge,wang2025can}.
\cite{li-etal-2025-generation} identified 4 application areas of \acrshort{llm}-as-a-Judge: evaluation, alignment, retrieval, and reasoning.
The work of \cite{zheng2023judging} showed that frontier \acrshort{llm} judges can match human preferences, demonstrating 85\% agreement using GPT-4.
This capability is confirmed in other works, even for domain-specific tasks \cite{szymanski2025limitations,huidrom-belz-2025-ask,wang2025can}.
In addition, less positional bias was observed with \acrshort{gpt}-4 compared to other frontier models of its time \cite{zheng2023judging}.
Scaling model size improves performance \cite{hoffmann2022training,NEURIPS2024_1cded4f9,dorigo2026codesign} but it has also been shown that this attribute can be task-dependent \cite{roberts-etal-2025-compute} and eventually saturates \cite{hoffmann2022training}.
The use of \acrshort{llm}-as-a-Judge has its limitations though, such as bias, sandbagging, and other challenges \cite{szymanski2025limitations,li-etal-2025-generation,adewumi-etal-2025-limitations,adewumi2025ai,shi2025judging}.

The requirement by \cite{act2023regulation} on the data format for \acrshort{dbp}s makes \acrshort{json} very suitable ~\cite{rfc8259}.
This is because it has (1) interoperability since it is language-agnostic and supports modern programming ecosystems, allowing seamless integration with machine learning frameworks~\cite{niu2015interoperability}, (2) machine-readability since its structured key–value representation enables efficient deterministic parsing~\cite{neubauer2025ai}, (3) structured representation of hierarchical data natively~\cite{lv2018survey}, and (4) searchability and efficient indexing, enabling flexible filtering and retrieval ~\cite{whairit2023jindex}.




\section{Conclusion}
\label{conclusion}

The \acrshort{bp} benchmark we have introduced provides a strong basis for evaluating the capability of \acrshort{ai} models in the novel \acrshort{dbp} conformance task.
As the \acrshort{eu} regulation on \acrshort{dbp} approaches, this work helps in preparing stakeholders on what to expect regarding \acrshort{ai} model predictions on \acrshort{dbp}s.
It showed that \textit{Thinking} models are more capable than other types of models and some \acrshort{slm}s have better performances than some \acrshort{llm}s.
The best-performing model showed signs of struggling on the test set, especially in correctly predicting conformant samples.
This weakness will be of more concern to battery (or \acrshort{dbp}) producers. 
Given the nature of \acrshort{dbp}s, which combine static and evolving multilingual and multimodal data, there are new tasks that are possible in the future, such as battery lifecycle reasoning and material extraction.
Since \textit{Thinking} models are designed to reason, they may still be \acrshort{sota} in such new tasks.



\section*{Ethics Statement}

While this work does not involve private, personal or sensitive data, we followed ethical standards in creating the dataset, including adding "synthetic" key to each data point.
We have also provided the prompts and links to the real pilot samples that were used in the generation of the dataset. 
We also publicly release the dataset and the metadata under a permissive license \acrshort{ccby}.

\section*{Societal Impact and Potential Harmful Consequences}
\label{bimpact}

Safety: There are no foreseeable situations in which this work or the dataset created under it can directly be used to harm, injure or kill people through its application, side effects, or potential misuse.

Environment: Although we used computing resources for our experiments, we kept this minimal by restricting experiments to in-context learning instead of pretraining or finetuning.
Furthermore, our research is not going to negatively impact the environment by promoting fossil fuel extraction, increasing societal consumption or producing substantial amounts of greenhouse gasses.

Bias and fairness: There are no relevant social biases in the dataset so it does not exacerbate bias against people of a certain gender, race, sexuality, or other protected characteristics.
However, the dataset has the limitations inherent in the real pilot samples, which are discussed in the next section. 
In addition, the dataset is balanced across the two classes: conformant and nonconformant.

Data documentation: Details of the dataset are communicated via a structured template of data statement in the appendix, in addition to the metadata that is publicly released with the dataset under the same license (\acrshort{ccby}).

Allowing access to research artifacts: We have made accessible the information required to understand our artifacts, including execution environments and hyper-parameters, to enable external scrutiny and auditing.

Essential elements for reproducibility: We have provided the information sufficient for the reproduction of results described, including descriptions of the packages, data, models and the evaluation times.

\section*{Limitations}
\label{limits}

We acknowledge this work has some limitations.
Our dataset \acrshort{bp} is based on pilot samples of the \acrshort{gba}, which only have some (10) of all the public fields of the description in the regulation by \cite{act2023regulation}.
When full \acrshort{dbp}s are in operation in 2027, it will be beneficial to update this dataset.
As a result, the  performances of the models on the full updated \acrshort{dbp}s may be different.
Furthermore, since the regulation by \cite{act2023regulation} applies only within the \acrshort{eu}, though it may be general enough, it may be that other countries or regions may define additional requirements for their own \acrshort{dbp}s.
For example, \acrshort{dbp}s may be multilingual - having dynamic inputs in different languages or multimodal - containing different modalities of input.
Our work assumed \acrshort{dbp}s to be only in English in this work.
In addition, our work focused on inconsistencies in \acrshort{dbp}s as the only basis for nonconformance.
However, full compliance will require that all the fields identified in \cite{act2023regulation} are present and valid for a \acrshort{dbp} to be classified as conformant.

\acrshort{gba} noted that their pilot \acrshort{dbp}s  had different scope of reporting across the consortia and the data gathering and score aggregation were exploratory.
Also, out of the many battery producers in the world, the pilot samples involve only a handful of partners or producers, thereby limiting the representation of battery producers in the \acrshort{dbp}s.
This currently biases the dataset to those producers.
Furthermore, though \acrshort{bp} is generated from real pilot samples, making it as realistic as possible, the data points are still synthetic and it is uncertain how actual performance of the models on real \acrshort{dbp}s may differ from the synthetic data. 



\section*{Appendix A - \acrshort{dbp} Samples}
\label{appen_dbps}

\subsection*{1. Conformant example with serial number: 0B5PEHT0A1082JE90136}

\{
  "source\_url": "https://www.globalbattery.org/battery-passport-mvp-pilots/pilot\-07\-h58fbc3dl0/",
  "pilot": {
    "year": 2024,
    "pilot\_id": "Pilot 07",
    "partners": [
      "CALB",
      "Smart",
      "PTL",
      "Circularise"
    ],
    "programme": "GBA Battery Passport 2024 pilots"
  },
  "battery\_information": {
    "battery\_serial\_number": "0B5PEHT0A1082JE90136",
    "battery\_chemistry": "LFP",
    "material\_flow\_aggregation": "individual battery",
    "battery\_model": "D300N97EA",
    "battery\_cell\_type": "prismatic",
    "tracing\_period\_start": "no info",
    "battery\_status": "Original",
    "number\_of\_cells\_per\_battery": 192,
    "tracing\_period\_end": "no info",
    "manufacturing\_date": "Jan 2024",
    "weight\_kg": 560,
    "data\_collection\_assured\_by": [
      "Circularise",
      "Shenzhen Precise Testing Technology"
    ],
    "ev\_oem": "Smart Automobile",
    "total\_energy\_kwh": 99.98,
    "country\_of\_ev\_production": "China",
    "energy\_density\_Wh\_per\_kg": 175,
    "traceability\_level": {
      "score": "1/3",
      "description": "basic"
    },
    "battery\_producer": "CALB",
    "rated\_capacity\_Ah": 279,
    "interoperability\_level": {
      "score": "1/3",
      "description": "basic"
    },
    "country\_of\_battery\_production": "China",
    "expected\_lifetime": 1500,
    "data\_verification\_assured\_by": "no applicable",
    "battery\_cell\_producer": "CALB",
    "voltage\_V\_min\_nominal\_max": [
      269,
      350,
      418
    ],
    "country\_of\_cell\_production": "China",
    "temperature\_range\_C\_min\_max": [
      -40,
      70
    ],
    "verification\_level": {
      "score": "0/3",
      "description": "low"
    },
    "comments\_and\_disclaimers": [
      "Technical parameters do not reflect the major focus of the GBA BP MVP pilots.",
      "Definitions of traceability, interoperability and verification can be found on the landing page.",
      "The tracing period data were mainly used for simplified mass balancing to determine suppliers proportions",
      "Battery number is limited to 20 symbols only."
    ]
  },
  "bill\_of\_materials": {
    "raw\_materials\_count": 10,
    "raw\_materials\_share\_percent": 100.0,
    "recycled\_materials\_share\_percent": 0.0,
    "materials\_traceability\_in\_pilot\_percent": "not applicable",
    "comments\_and\_disclaimers": [
      "GBA members and partners",
      "1) As per the conditions of the pilots the value of the physical weight and supply shares could have been smoothed or rounded",
      "2) Materials tracking or tracing did not automatically mean sharing of ESG data",
      "3) The requirements to hazardous components are not yet fully defined"
    ]
  },
  "materials": {
    "primary\_traced\_material": {
      "material": "not applicable",
      "physical\_amount\_kg": "",
      "traced\_amount\_kg": "",
      "company\_share\_percent": 100.0,
      "country": "",
      "esg\_performance": "reported",
      "other\_companies\_share\_percent": "",
      "other\_companies\_traceability": ""
    }
  },
  "esg\_performance\_summary\_raw\_table": "ESG performance is disclosed across mining, refining and overall clusters.\\n\\nCompany counts:\\n- Mining: 1\\n- Refining: 1\\n- Overall: 2\\n\\nQuantitative Issue 01: Greenhouse Gas (GHG)\\n- PMA and HMA allocations are shown.\\n- GHG (PMA) and GHG (HMA) values include fully reported intensity figures.\\n- Primary data share is reported as 100\% for both mining and refining.\\n\\n- Child labour due diligence\\n- Environmental and human-rights due diligence\\n- Biodiversity due diligence\\n- Circular design\\n- Forced labour due diligence\\n- Indigenous peoples\\u2019 rights\\n\\nThe tables include:\\n, Evidence types (data proofs): external audits, standard-based checks, self-reported data and validated documents,\\n- Verified report counts and total report counts.\\n\\nAdditional secured ESG score tables show high overall performance with low n/a rates and strong primary data presence.\\n",
  "synthetic": true
\}

\subsection*{2. Nonconformant example with wrong array length and linked values: 0B5PEHT0A1082JE9049E}

\{
  "source\_url": "https://www.globalbattery.org/battery-passport-mvp-pilots/pilot-07-h58fbc3dl0/",
  "pilot": {
    "year": 2024,
    "pilot\_id": "Pilot 07",
    "partners": [
      "CALB",
      "Smart",
      "PTL",
      "Circularise"
    ],
    "programme": "GBA Battery Passport 2024 pilots"
  },
  "battery\_information": {
    "battery\_serial\_number": "0B5PEHT0A1082JE9049E",
    "battery\_chemistry": "LFP",
    "material\_flow\_aggregation": "individual battery",
    "battery\_model": "D300N97EA",
    "battery\_cell\_type": "prismatic",
    "tracing\_period\_start": "no info",
    "battery\_status": "Original",
    "number\_of\_cells\_per\_battery": 192,
    "tracing\_period\_end": "no info",
    "manufacturing\_date": "Jan 2024",
    "weight\_kg": 560,
    "data\_collection\_assured\_by": [
      "Circularise",
      "Shenzhen Precise Testing Technology"
    ],
    "ev\_oem": "Smart Automobile",
    "total\_energy\_kwh": 100,
    "country\_of\_ev\_production": "China",
    "energy\_density\_Wh\_per\_kg": 175,
    "traceability\_level": {
      "score": "1/3",
      "description": "basic"
    },
    "battery\_producer": "CALB",
    "rated\_capacity\_Ah": 279,
    "interoperability\_level": {
      "score": "1/3",
      "description": "basic"
    },
    "country\_of\_battery\_production": "China",
    "expected\_lifetime": 1500,
    "data\_verification\_assured\_by": "no applicable",
    "battery\_cell\_producer": "CALB",
    "voltage\_V\_min\_nominal\_max": [
      269,
      350
    ],
    "country\_of\_cell\_production": "China",
    "temperature\_range\_C\_min\_max": [
      -40,
      70
    ],
    "verification\_level": {
      "score": "0/3",
      "description": "low"
    },
    "comments\_and\_disclaimers": [
      "Technical parameters do not reflect the major focus of the GBA BP MVP pilots.",
      "Definitions of traceability, interoperability and verification can be found on the landing page.",
      "The tracing period data were mainly used for simplified mass balancing to determine suppliers proportions",
      "Battery number is limited to 20 symbols only."
    ]
  },
  "bill\_of\_materials": {
    "raw\_materials\_count": 10,
    "raw\_materials\_share\_percent": 100.0,
    "recycled\_materials\_share\_percent": 0.0,
    "materials\_traceability\_in\_pilot\_percent": "not applicable",
    "comments\_and\_disclaimers": [
      "GBA members and partners",
      "1) As per the conditions of the pilots the value of the physical weight and supply shares could have been smoothed or rounded",
      "2) Materials tracking or tracing did not automatically mean sharing of ESG data",
      "3) The requirements to hazardous components are not yet fully defined"
    ]
  },
  "materials": {
    "primary\_traced\_material": {
      "material": "not applicable",
      "physical\_amount\_kg": "",
      "traced\_amount\_kg": "",
      "company\_share\_percent": 70.0,
      "country": "",
      "esg\_performance": "reported",
      "other\_companies\_share\_percent": 50.0,
      "other\_companies\_traceability": ""
    }
  },
  "esg\_performance\_summary\_raw\_table": "ESG performance is disclosed across mining, refining and overall clusters.\\n\\nCompany counts:\\n- Mining: 1\\n- Refining: 1\\n- Overall: 2\\n\\nQuantitative Issue 01: Greenhouse Gas (GHG)\\n- PMA and HMA allocations are shown.\\n- GHG (PMA) and GHG (HMA) values include fully reported intensity figures.\\n- Primary data share is reported as 100\% for both mining and refining.\\n\\n- Child labour due diligence\\n- Environmental and human-rights due diligence\\n- Biodiversity due diligence\\n- Circular design\\n- Forced labour due diligence\\n- Indigenous peoples\\u2019 rights\\n\\nThe tables include:\\n, Evidence types (data proofs): external audits, standard-based checks, self-reported data and validated documents,\\n- Verified report counts and total report counts.\\n\\nAdditional secured ESG score tables show high overall performance with low n/a rates and strong primary data presence.\\n",
  "synthetic": true
\}

\section*{Appendix B - Prompts}
\label{append_prompts}

\subsection*{Retrieval Prompt}
\begin{quote}
    Create a new JSON file of the content of this website on digital battery passport for me to download: \textit{link}
\end{quote}

\subsubsection*{Data-Metadata Generation Prompt}
\begin{quote}
Generate 2000 unique JSON samples from this conformant digital battery passport, where half (1000) are conformant and the other half are nonconformant.
Add only 1 additional field at the end of each of the 2000 samples indicating them as synthetic (true).
Zip all the conformant samples in one downloadable file and the other nonconformant samples in another zip.
Do not bias the samples in any way, such as using terms related to conformant in the filename or serial number, which should be their filenames.
The 1000 conformant samples must preserve quality while being unique in battery\_serial\_number and be such that a fourth (250 samples) have exactly the same values in all other fields, another fourth should have different values only in 1 field, another fourth should have different values only in 2 fields, and the last fourth should have different values only in 3 fields.
The fields can be from any of these three: total\_energy\_kwh, expected\_lifetime, and voltage\_V\_min\_nominal\_max.

Also, the 1000 nonconformant samples should be diverse such that about sixth of them should have only 1 internal inconsistency, another sixth should have only 2 internal inconsistencies, another sixth should have only 3 internal inconsistencies, another sixth should have only 4  internal inconsistencies, another sixth should have only 5 internal inconsistencies, and the last sixth should have only 6 internal inconsistencies.
The internal inconsistencies can be any of these six: (1) inconsistencies in linked values, (2) unrealistic data entries, (3) inconsistency in physical and traced amounts, (4) conflicting dates, (5) invalid codes, and (6) wrong length of arrays.
All the 2000 samples that are generated and their details of diversity or inconsistencies should be documented in a downloadable metadata Excel file, and must include 2 key columns: fields\_changed\_from\_base (to list the fields that changed) and num\_changed (to put the total fields changed for each sample).
Follow all these instructions correctly and ensure there are no errors when documenting the details in the metadata file.
\end{quote}

\subsection{\acrshort{llm}-as-a-Judge Prompt}

\begin{quote}

    Compare each JSON file in these zip files with their original JSON \textit{filename}.
    Document all the differences of each JSON (including the synthetic field) to the original JSON file in the provided metadata Excel file by adding 2 key columns: validation\_fields\_changed\_from\_base (to list the fields that changed separated by semi-colon) and validation\_num\_changed (to put the total fields changed for each sample).
    Count all differences with the field voltage\_V\_min\_nominal\_max as one for each sample.
    Then add a 3rd column (validation - true or false) to the Excel file to compare the 2 columns fields\_changed\_from\_base and validation\_fields\_changed\_from\_base in each row, where it is true if they contain exactly the same fields but false otherwise and use this to calculate the overall accuracy by dividing the count of true instances in the validation column by the total rows.
\end{quote}


\section*{Appendix C - Models}
\label{appen_models}
While some of the models can overlap into different categories, we have tried to classify them into the following (as represented in the results - Table \ref{full_results}), based on their sizes and training approach, as described in some model cards.
We provide model cards in the next subsection, which gives more details and hyper-parameters, where relevant.

\textbf{\textit{General}} models:
Here we include \acrfull{moe} and dense models since they tend to have large total parameters.
\acrshort{moe}s combine specialized sub-networks forming a large number of parameters.
They use a router at inference to select the active parameters of the relevant expert sub-networks for answers.
Meanwhile, traditional dense models have large number of parameters and use all the parameters during inference.
Dense models are simpler to implement than \acrshort{moe}.

\textbf{\textit{Thinking}} models: They are designed to spend more time and effort, through complex algorithms, before answering.
Their reasoning usually involves multi-step logic and chain of thought, which can make them more expensive.

\textbf{\textit{Instruct}} models: They are finetuned to follow specific human instructions, thereby aligning them with certain human values.
They do not necessarily have the reasoning depth of \textit{Thinking} models.

\textbf{\textit{Pro}} models: They are generally designed for tough problems that require deeper reasoning and can take longer to solve.

\textbf{\textit{\acrshort{slm}}}:
They can take the form of any of the earlier models but they have far fewer parameters and, therefore, lesser memorization capability and broad knowledge.

\subsection*{Hyper-parameters}
We use default hyper-parameters, as designed and recommended by the platforms.

\acrshort{gpt} family: \textit{Reasoning.effort=none},  \textit{Verbosity=medium} (https://developers.openai.com/api/docs/guides/latest-model)

Gemini-3 family: \textit{Thinking-level=high}, \textit{Temperature=1.0} (https://ai.google.dev/gemini-api/docs/gemini-3)

Claude family: \textit{Thinking=adaptive}, \textit{Temperature=1.0} (https://docs.aimlapi.com/api-references/text-models-llm/anthropic/claude-4.6-sonnet)

HuggingChat: We provide links to the model cards from HuggingChat.

\begin{enumerate}
    \item cogito-671b-v2.1 - https://huggingface.co/deepcogito/cogito-671b-v2.1
    \item gpt-oss-120b - https://huggingface.co/openai/gpt-oss-120b
    \item Qwen3.5-397B-A17B - https://huggingface.co/Qwen/Qwen3.5-397B-A17B
    \item Qwen3.5-122B-A10B - https://huggingface.co/Qwen/Qwen3.5-122B-A10B
    \item MiniMax-M2.1 - https://huggingface.co/MiniMaxAI/MiniMax-M2.1
    \item WizardLM-2-8x22B - https://huggingface.co/alpindale/WizardLM-2-8x22B
    \item ERNIE-4.5-VL-424B-A47B-Base-PT - https://huggingface.co/baidu/ERNIE-4.5-VL-424B-A47B-Base-PT
    \item Kimi-K2-Thinking - https://huggingface.co/moonshotai/Kimi-K2-Thinking
    \item Llama-4-Scout-17B-16E-Instruct - https://huggingface.co/meta-llama/Llama-4-Scout-17B-16E-Instruct
    \item Qwen3-4B-Instruct-2507 - https://huggingface.co/Qwen/Qwen3-4B-Instruct-2507
    \item gemma-3-27b-it - https://huggingface.co/google/gemma-3-27b-it
\end{enumerate}


\subsection*{Pricing}

\acrshort{gpt} family: https://developers.openai.com/api/docs/models/gpt-5.4

Claude family: https://platform.claude.com/docs/en/about-claude/models/overview

Gemini-3 family: https://ai.google.dev/gemini-api/docs/gemini-3

HuggingChat: https://huggingface.co/pricing 

\section*{Appendix D - \acrshort{bp} Data Statement}
\label{bp_datastate}

\subsection*{Curation Rationale}
\acrshort{bp} was created based on the  need to have high-quality \acrfull{dbp} dataset as the \acrshort{eu} regulation on \acrshort{dbp} comes into effect and no dataset currently exists.
The dataset is designed for conformant or nonconformant binary classification task in the first instance.
It is also applicable for other reasoning tasks related to batteries or \acrshort{dbp}s.

\subsection*{Generation process}
The dataset was synthetically generated from real pilot samples by the \acrfull{gba} using \acrshort{gpt}-5.1 \textit{Thinking (Standard)} with default hyper-parameters of the chat \acrfull{ui} of ChatGPT from OpenAI.

The steps are:
(1) we manually loaded each \acrshort{dbp} pilot sample through its \acrshort{url} and establish if it is valid (i.e. contains all the main 10 pieces of information, resulting in 6 valid samples out of 10, (2) we fed the \acrshort{url} to ChatGPT-5.1 \textit{Pro} for automatic document retrieval in JSON format, (3) we
performed \acrfull{qc} by assessing if all the relevant elements of each pilot sample are present in the retrieved
JSON, resulting in 3 that needed manual correction, (4) we used ChatGPT-5.1 \textit{Thinking (Standard)} to
automatically generate (on average time of 47 seconds) 2,000 synthetic samples and their metadata
from each of the 6 pilot samples (where 1,000 are conformant like the original pilot while the other 1,000
are nonconformant), (5) we evaluated all the 12,000 samples and their metadata automatically, using
ChatGPT-5.0 \textit{Thinking (Standard)} giving an \textbf{accuracy of 99.68\%}, before a random sample is also humanly evaluated, and (6) we
randomized and split the data into training, validation and test sets in the ratio 80:10:10.

\subsection*{Language Variety}
The dataset is in English (en-US and international English.)

\subsection*{Speaker demographic}
Not applicable.

\subsection*{Annotator demographic}
Not applicable.

\subsection*{Dataset structure}
The fields in the dataset include the following:

\begin{enumerate}
    \item source\_url: source of the base pilot sample.
    \item year: year of production.
    \item pilot\_id: base pilot number.
    \item partners: partners involved in the battery production.
    \item battery\_information: key information about the battery, e.g. serial number and battery chemistry.
    \item battery\_producer: battery manufacturer.
    \item rated\_capacity\_Ah: rated capacity in Ah.
    \item interoperability\_level: description of interoperability.
    \item country\_of\_battery\_production: country where the battery is produced.
    \item expected\_lifetime: expected\_lifetime in cycles.
    \item voltage\_V\_min\_nominal\_max: array value of minimum, nominal and maximum voltage.
    \item temperature\_range\_C\_min\_max: range of temperature from minimum to maximum.
    \item comments\_and\_disclaimers: \acrshort{gba} comments and disclaimer.
    \item bill\_of\_materials: more details about the battery materials.
    \item materials: information about primary and secondary traced materials.
    \item esg\_performance\_summary\_raw\_table: Performance summary of materials sourced according to Environmental, Social, and Governance (ESG) standards.

\end{enumerate}

The dataset is split into 3 sets:

\begin{enumerate}
    \item Training set of 9,600 samples
    \item Validation set of 1,200 samples
    \item Test set of 1,200 samples
\end{enumerate}

\subsection*{Benchmarking}
The dataset was benchmarked across 22 \acrshort{ai} models in zero-shot inference (among other tests) against a random baseline of 0.49\% (0.04) F1 (and confidence interval at 95\% level).
The best-performing model was \acrshort{gpt}-5.4 \textit{Thinking (Standard)} with 0.98\% (0.03) and 0.71\% (0.22) F1 for the validation and test sets, respectively.

\subsection*{Others}
The dataset is licenced under \acrshort{ccby}.
The dataset is limited in that the pilot samples used to create it involve only a handful of partners or producers, thereby limiting
representation of battery producers.
This currently biases the dataset to those producers.

\begin{ack}
This work is partly supported by the Wallenberg AI, Autonomous Systems and Software Program (WASP), funded by Knut and Alice Wallenberg Foundations.
\end{ack}


{
\small

\bibliographystyle{IEEEtran}
\bibliography{ref}

}


\appendix




\newpage
\section*{}

\end{document}